\documentclass[runningheads]{llncs}
\usepackage{graphicx}
\usepackage{amsmath,amssymb} 
\usepackage{color}
\usepackage{cite}
\usepackage[pagebackref=true,breaklinks=true,letterpaper=true,colorlinks,bookmarks=false]{hyperref}
\begin{document}
\newcommand{\etal}{\emph{et al.}}
\newcommand{\eg}{\emph{e.g.}}
\newcommand{\ie}{\emph{i.e.}}
\newcommand{\etc}{\emph{etc}}

\title{Pose Guided Human Video Generation} 
\titlerunning{Pose Guided Human Video Generation}
\author{Ceyuan Yang\inst{1} \and
Zhe Wang\inst{2} \and
Xinge Zhu\inst{1} \and
Chen Huang \inst{3} \and \\
Jianping Shi \inst{2} \and
Dahua Lin \inst{1}}
\authorrunning{C. Yang, Z. Wang, X. Zhu, C. Huang, J. Shi, D. Lin \etal}

\institute{CUHK-SenseTime Joint Lab, CUHK, Hong Kong S.A.R. \and
SenseTime Research, Beijing, China  \and 
Carnegie Mellon University \\ \email{yangceyuan@sensetime.com}
}
\maketitle              

\begin{abstract}
Due to the emergence of Generative Adversarial Networks, video synthesis has witnessed exceptional breakthroughs. However, existing methods lack a proper representation to explicitly control the dynamics in videos. Human pose, on the other hand, can represent motion patterns intrinsically and interpretably, and impose the geometric constraints regardless of appearance. In this paper, we propose a pose guided method to synthesize human videos in a disentangled way: \emph{plausible motion prediction} and \emph{coherent appearance generation}. In the first stage, a Pose Sequence Generative Adversarial Network (PSGAN) learns in an adversarial manner to yield pose sequences conditioned on the class label. In the second stage, a Semantic Consistent Generative Adversarial Network (SCGAN) generates video frames from the poses while preserving coherent appearances in the input image. By enforcing semantic consistency between the generated and ground-truth poses at a high feature level, our SCGAN is robust to noisy or abnormal poses. Extensive experiments on both human action and human face datasets manifest the superiority of the proposed method over other state-of-the-arts.
\end{abstract}

\section{Introduction}
\label{Intro}

With the emergence of deep convolution networks, a large amount of generative models have been proposed to synthesize images, such as Variational Auto-Encoders~\cite{kingma2014auto} and Generative Adversarial Networks~\cite{goodfellow2014generative}. Meanwhile, video generation and video prediction tasks~\cite{srivastava2015unsupervised, finn2016unsupervised, mathieu2015deep, oh2015action} have found big progress as well. Among them, the task of human video generation attracts increasing attention lately. One reason is that human video synthesis allows for many human-centric applications like avatar animation. On the other hand, generation of human videos/frames can act as a data augmentation method which largely relieves the burden of manual annotations. This will speed up the development of a wide range of video understanding tasks such as action recognition.

Human video generation is a non-trivial problem itself. Unlike static image synthesis, the human video generation task not only needs to take care of the temporal smoothness constraint but also the uncertainty of human motions. Therefore, a proper representation of human pose and its dynamics plays an important role in the considered problem. Recent works attempted to model video dynamics separately from appearances. For instance, in~\cite{denton2017unsupervised} each frame is factorized into a stationary part and a temporally varying component. Vondrick~\etal~\cite{vondrick2016generating} untangled the foreground scene dynamics from background with a two-stream generative model. Saito~\etal~\cite{saito2017temporal} generated a set of latent variables (each corresponds to an image frame) and learned to transform them into a video. Tulyakov~\etal~\cite{tulyakov2017mocogan} generated a sequence of video frames from a sequence of random vectors, each consists of a content part and a motion part. All these methods show the promise of separate modeling of motion dynamics and appearances. However, motions cannot be controlled explicitly in these methods. The motion code is usually sampled from a random latent space, with no physical meaning about the targeted motion pattern.

Here we argue that for human video generation, to model human dynamics effectively and control motions explicitly, the motion representation should be interpretable and accessible. Inspired by the action recognition literature~\cite{hodgins1998perception, yan2018spatial, du2015hierarchical, vemulapalli2014human},  human body skeletons are favorable in that they characterize the geometric body configuration regardless of the appearance difference, and their dynamics capture motion patterns interpretably. It is also worth noting that human skeletons can be easily obtained by many state-of-the-art human pose estimators (\eg~\cite{cao2017realtime}).

\begin{figure}[t]
\centering
\includegraphics[width=0.9\textwidth]{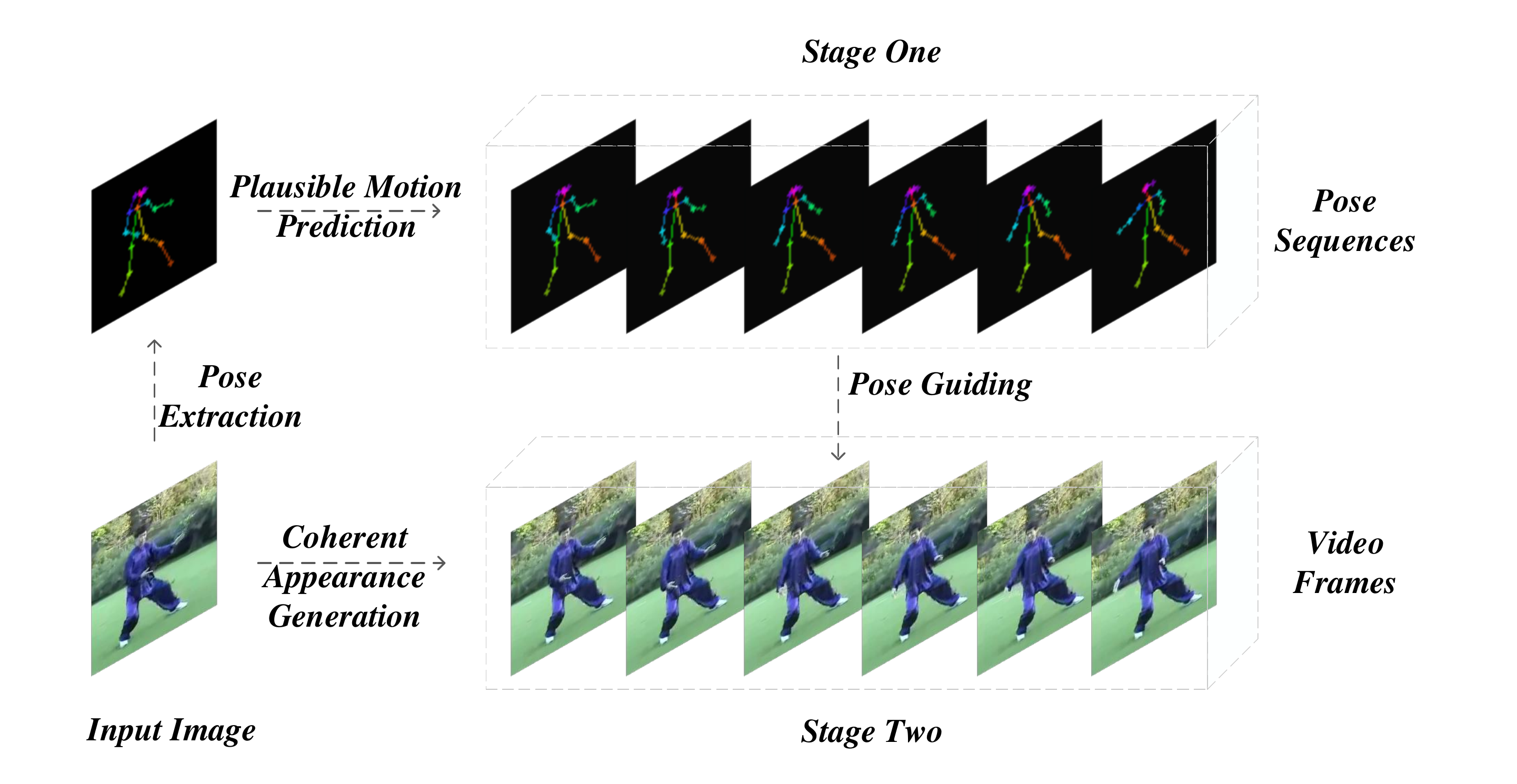}
\caption{The framework of our method. In the first stage, we extract the corresponding pose for an input image and feed the pose into our PSGAN to generate a pose sequence. In the second stage, SCGAN synthesizes photo-realistic video frames given the generated poses and input image}
\label{fig:PIPELINE}
\end{figure}

Therefore, we propose a pose guided method to synthesize human videos. The method consists of two stages: \emph{plausible motion prediction} and \emph{coherent appearance generation}, generating the pose dynamics and corresponding human appearances separately. In the first stage,  human pose is used to model various motion patterns. The Pose Sequence Generative Adversarial Network (PSGAN) is proposed to learn such patterns explicitly, conditioned on different action labels. In the second stage, a Semantic Consistent Generative Adversarial Network (SCGAN) is proposed to generate video frames given the generated pose sequence in the first stage and input image. Meanwhile, the semantic consistency between the generated and ground-truth poses is also enforced at a high feature level, in order to alleviate the influence of some noisy or abnormal poses. Experiments will show the efficacy and robustness of our method when generating a wide variety of human action and facial expression videos. Fig.~\ref{fig:PIPELINE} illustrates the overall framework. We summarize the major contributions of as follows: 
\begin{itemize}
\item [$\bullet$]We propose a Pose Sequence Generative Adversarial Network (PSGAN) for plausible motion prediction based on human pose, which allows us to model the dynamics of human motion explicitly.
\item [$\bullet$]Semantic Consistent Generative Adversarial Network (SCGAN) is designed to synthesize coherent video frames given the generated pose and input image, with an effective mechanism for handling abnormal poses.
\item [$\bullet$]Qualitative and quantitative results on human action and facial expression datasets show the superiority of the proposed method over prior arts. The controlled experiments also show our flexibility to manipulate human motions as well as appearances. Codes will be made publicly available.
\end{itemize}

\section{Related work}
\label{Related}
Deep generative models have been extensively studied to synthesize natural images, typically using Variational Auto-Encoders (VAEs)~\cite{kingma2014auto} and Generative Adversarial Networks (GANs)~\cite{goodfellow2014generative}. Many follow-up works aim to improve the training of GANs~\cite{goodfellow2014generative} and thus enhance the generated image quality. The authors of~\cite{radford2015unsupervised} noted that the uncertainty of data distribution is likely to cause model collapse, and proposed to use convolutional networks to stabilize training. The works in~\cite{arjovsky2017wasserstein,gulrajani2017improved} handle the problem of GAN training instability as well.

Another direction is to explore image generation in a conditioned manner. The pioneer work in~\cite{mirza2014conditional} proposes a conditional GAN to generate images controlled by class labels or attributes. More recently, Ma~\etal~\cite{ma2017pose} proposed a pose guided person generation network to synthesize person images in arbitrary new poses. StackGAN~\cite{zhang2017stackgan} is able to generate photo-realistic images from some text descriptions. The works in~\cite{zhu2017unpaired, yi2017dualgan, kim2017learning} further generalize to learn to translate data from one domain to another in an unsupervised fashion. StarGAN~\cite{choi2017stargan} even allows to perform image-to-image translation among multiple domains with only a single model. Our proposed PSGAN and SCGAN are also designed conditional, generating a human pose sequence given an action label and then generating video frames given the pose sequence and input image. Our two conditional models are able to generate a continuous human video at one time rather than static images only. The SCGAN can also alleviate the impact of abnormal poses by learning semantic pose representations.

The task of video generation is intrinsically a much more challenging one than the image synthesis task, due to the requirement of foreground dynamics modeling and temporal smoothness constraint. During the past few years, it is with the access to powerful GPUs and the advent of deep convolution networks that video generation and video prediction~\cite{srivastava2015unsupervised, finn2016unsupervised, mathieu2015deep, oh2015action} have gained large momentum as well. As an example, one GAN model with a spatio-temporal convolutional architecture is proposed in~\cite{vondrick2016generating} to model the foreground scene dynamics in videos. Tulyakov~\etal~\cite{tulyakov2017mocogan} also decomposed motion and content for video generation. In~\cite{liang2017dual} future-frame predictions are made consistent with the pixel-wise flows in videos through a dual learning mechanism. Other works introduce recurrent networks into video generation (\eg~\cite{fragkiadaki2015recurrent,zhou2016learning}). In line with these works, our method separately models the motion and appearance as well, using the PSGAN and SCGAN respectively. This enables us to control the motion patterns explicitly and interpretably, which to the best of our knowledge, is the first attempt in human video generation.

\section{Methodology}
\label{Method}
\subsection{Framework Overview}
\label{Overview}
Given an input image of human body or face and a target action class (\eg,~\emph{Skip}, \emph{TaiChi}, \emph{Jump}), our goal is to synthesize a video of human action or facial expression belonging to the target category and starting with the input image. We wish to explicitly control the motion patterns in the generated video while maintaining appearance coherence with the input. We here propose to generate human videos in a disentangled way: \emph{plausible motion prediction} and \emph{coherent appearance generation}. Fig.~\ref{fig:PIPELINE} illustrates the overall framework of our method. 

Similar to the action recognition literature~\cite{hodgins1998perception, yan2018spatial, du2015hierarchical, vemulapalli2014human}, we use the human skeletons or human pose for representations of motion dynamics. Our method consists of two stages. In the first stage, we extract the pose from input image and the Pose Sequence GAN (PSGAN) is proposed to generate a temporally smooth pose sequence conditioned on the pose of input image and the target action class. In the second stage, we focus on appearance modeling and propose a Semantic Consistent GAN (SCGAN) to generate realistic and coherent video frames conditioned on the input image and pose sequence from stage one. The impact of noisy/abnormal poses is alleviated by maintaining semantic consistency between generated and ground-truth poses in high-level representation space. Details will be elaborated in the following sections.
\begin{figure}[t]
\centering
\includegraphics[width=0.9\textwidth]{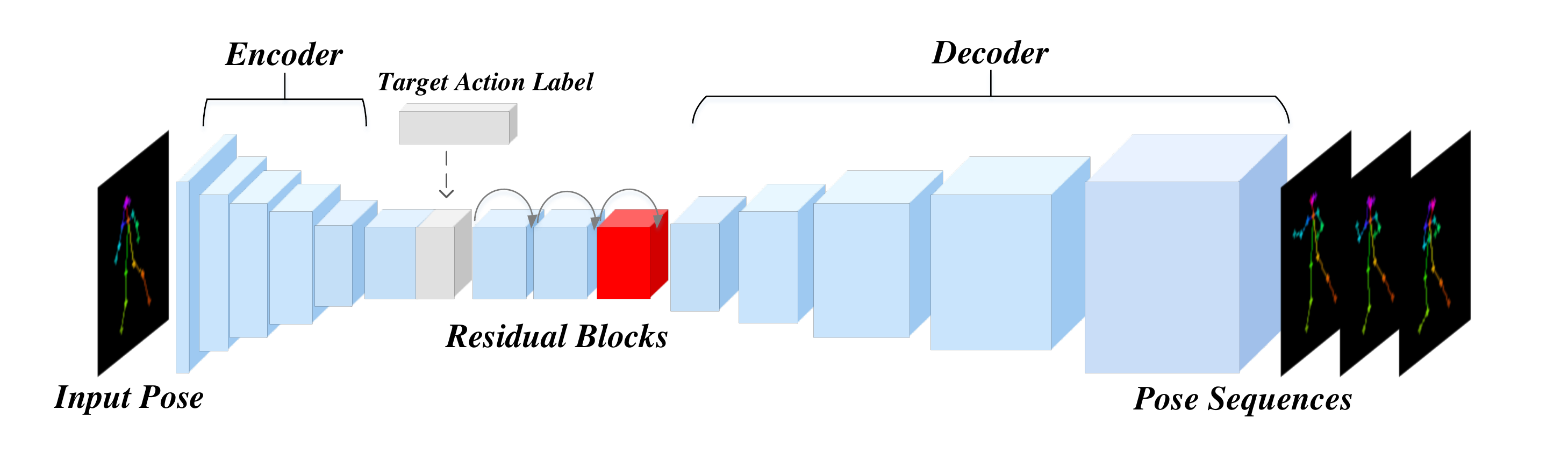}
\caption{Network architecture of our Pose Sequence GAN (PSGAN). PSGAN takes the input pose and target action label as input, and synthesizes pose sequences in an encoder-decoder manner. After the last residual block (red), the feature map is extended with a time dimension and then fed into the decoder which is composed of a series of fractionally-strided spatio-temporal convolution layers}
\label{fig:PSGAN}
\end{figure}

\subsection{Plausible Motion Prediction}
\label{One}
In the first stage, the human pose extracted from input image together with the target action label is fed into our PSGAN to generate a sequence of poses. Obviously this is an ill-posed one-to-many problem with infinite possibilities. Our PSGAN learns from example pose sequences in the training set to mimic plausible motion patterns. Thus our learning objective is the modeling of rich motion patterns instead of the precise pose coordinates.

\subsubsection{Pose extraction.} 
\label{Posemaker}
To extract the initial pose from input image, a state-of-the-art pose estimator in~\cite{cao2017realtime} is adopted to produce the coordinates of 18 key points. The pose is encoded by 18 heatmaps rather than coordinate vectors of the key points. Each heatmap is filled with 1 within a radius of 4 pixels around the corresponding key point and 0 elsewhere. Consequently, pose is actually represented as a $C=18$ channel tensor. In this way, there is no need to learn how to map the key points into body part locations.

\subsubsection{Pose Sequence GAN.} 
\label{PSGAN}
Given the initial pose and the target action label, our PSGAN aims to synthesize a meaningful pose sequence at one time. As shown in Fig.~\ref{fig:PSGAN}, PSGAN adopts an encoder-decoder architecture. The $C \times W \times H$-sized pose is first encoded through several convolutional layers. The target action label is also inputted in the form of $n$-dimensional one-hot vector where $n$ denotes the number of action types. After a few residual blocks, the two signals are embedded into common feature maps in the latent space. These feature maps will finally go through the decoder with a extended time dimension. The output is a $C \times T \times W \times H$-sized tensor via a series of fractionally-strided spatio-temporal convolution layers, where $T$ denotes the number of time steps in the sequence. For the sake of better temporal modeling, the LSTM module~\cite{hochreiter1997long} is integrated into our network as well. In summary, we define a generator $G$ that transforms an input pose $p$ into a pose sequence $\hat{P}$ conditioned on the target action label $a$,~\ie,~$\emph{G}(\emph{p}, \emph{a}) \Rightarrow \hat{P}$. We train the generator $G$ in an adversarial way - it competes with a discriminator $D$ as the PatchGAN~\cite{isola2017image} which classifies local patches from the ground-truth and generated poses as real or fake.

\subsubsection{LSTM embedding.} 
\label{LSTM}
As is mentioned above, the decoder outputs a $C$$\times$$T$$\times$$W$$\times$$H$ tensor. It can be regarded as $T$ tensors with size $C \times W \times H$, all of which are fed into a one-layer LSTM module for temporal pose modeling. Our experiments will demonstrate that the LSTM module stabilizes training and improves the quality of generated pose sequences. 
\subsubsection{Objective function.}
As in \cite{goodfellow2014generative}, the objective functions of our PSGAN can be formulated as follows:
\begin{align}
\mathcal{L}_{adv}^{D} &= \mathbb{E}_{P}[\log D(P)] + \mathbb{E}_{p,a}[\log (1 - D(G(p,a)))], \\
\mathcal{L}_{adv}^{G} &= \mathbb{E}_{p,a}[\log(D(G(p,a)))],
\end{align}
where $\mathcal{L}_{adv}^{D}$ and $\mathcal{L}_{adv}^{G}$ denote the adversarial loss terms for discriminator $D$ and generator $G$, respectively. The discriminator $D$ aims to distinguish between the generated pose sequence \emph{G}(\emph{p},\emph{a}) and ground-truth \emph{P}. Moreover, we find adding a reconstruction loss term can stabilize the training process. The reconstruction loss is given below:
\begin{align}
\label{maskl1}
& \mathcal{L}_{rec} = \lambda_{rec} || (P - G(p,a)) \odot (\alpha M + 1) ||_{1},
\end{align}
where $M$ denotes the mask for each of the key point heatmap, $\odot$ denotes pixels-wise multiplication and $\lambda_{rec}$ is the weight for this $L1$ loss. The introduction of mask $M$ is due to the heatmap sparsity and imbalance of each key point, which makes learning difficult. We use the ground-truth $P$ as mask $M$ to mask out the small region around each key point for loss computation. Note when the scaling factor $\alpha=0$, this loss term is reduced to unweighted $L1$ loss.

\begin{figure}[!b]
\centering
\includegraphics[scale=0.8]{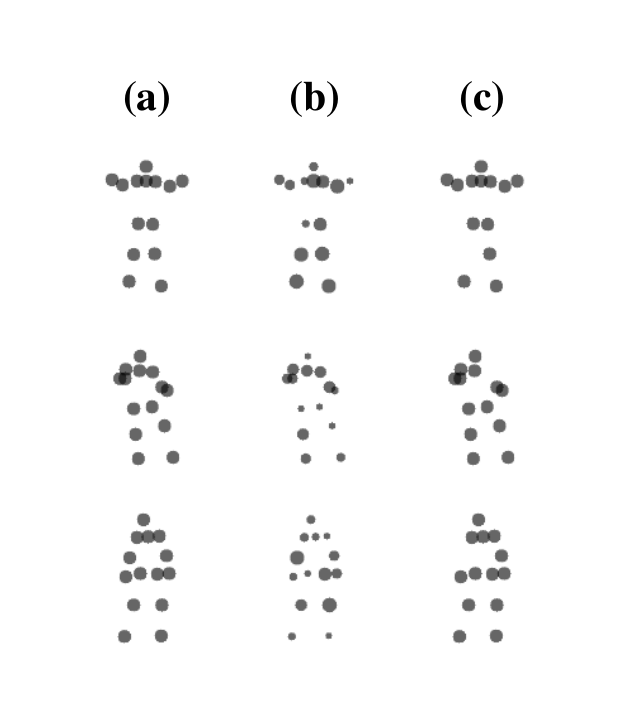}
\caption{Examples of abnormal poses. (a-c) show the ground-truth pose, generated poses with bigger/smaller key point responses, and with missing key points respectively}
\label{fig:AP}

\end{figure}

\subsubsection{Abnormal poses.} 
\label{AP}
Fig.~\ref{fig:AP} shows some bad pose generation results where some key points seem bigger/smaller (b) than the ground-truth (a), or some key points seem missing (c) because of their weak responses. We call such cases as abnormal poses. For human beings however, abnormal poses might look weird at the first glance, but would hardly prevent us from imagining what the ``true'' pose is. This requires our network to grasp the semantic implication of human pose, and to alleviate the influence of small numerical differences.

\subsection{Coherent Appearance Generation}

\begin{figure}[t]
\centering
\includegraphics[width=0.9\textwidth]{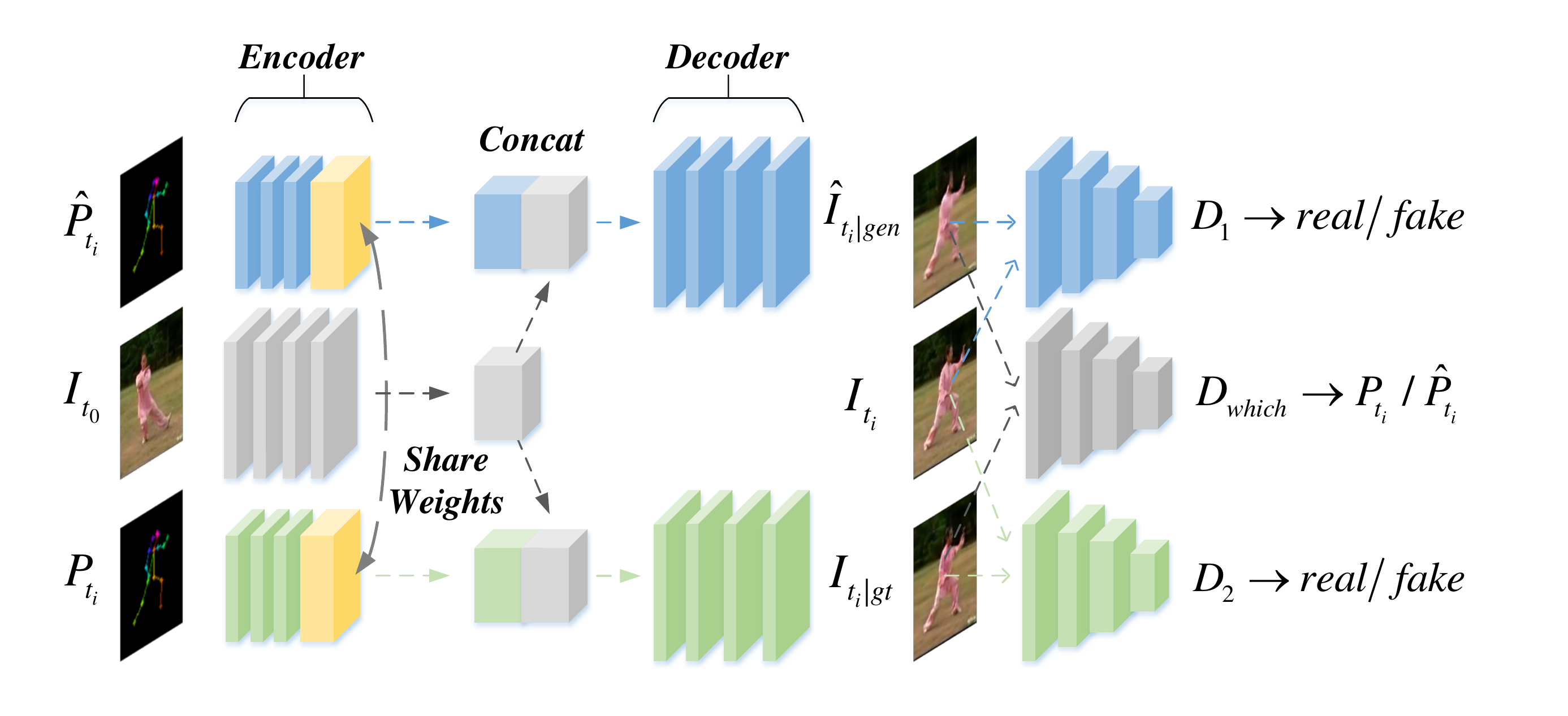}
\caption{Network architecture of our SCGAN in the second stage, where $\hat{P}_{t_{i}},P_{t_{i}},I_{t_{i}}$ respectively denote the pose generated by our method in stage one, ground-truth pose and the original image. Our generator has an encoder-decoder architecture and generates video frames conditioned on human poses $P$ and the input image $I_{t_0}$. Discriminators $D_{1}$ and $D_{2}$ aim to distinguish whether the generated images are real, while $D_{which}$ aims to tell which pose the frame is generated from}
\label{fig:APPGAN}
\end{figure}

\label{Ill}
In the second stage, we aim to synthesize coherent video frames conditioned on the input image as well as the pose sequence from stage one. Since the noisy or abnormal poses will affect image generation in this stage, those methods (\eg~\cite{ma2017pose}) to directly generate images from pose input may be unstable or even fail. Therefore, we propose a Semantic Consistent GAN (SCGAN) to impose semantic consistency between generated pose and ground-truth at a high feature level. By only imposing the consistency at high feature level, SCGAN can be robust to noisy pose inputs.

\subsubsection{Conditional generation.} Our conditional image generation process is actually similar to that in recent work~\cite{ma2017pose} which can generate person images controlled by pose. However, we have a major difference with this work: in~\cite{ma2017pose} images are generated in two stages by synthesizing a coarse image first and then refining it; while our SCGAN generates results in one step, for all video frames over time at once. Specifically, given the input image $I_{t_0}$ at time $t_0$ and the target pose $P_{t_i}$ at time $t_i$, our generator $G(I_{t_{0}}, P_{t_{i}}) \Rightarrow \hat{I}_{t_{i}}$ is supposed to generate image $\hat{I}_{t_i}$ to keep the same appearance in $I_{t_0}$ but on the new pose $P_{t_{i}}$. We design the discriminator $D$ again to tell real image from fake to improve the image generation quality.

\subsubsection{Semantic consistency.}
As discussed before, the noisy or abnormal pose prediction $\hat{P}_{t_{i}}$ from the first stage will affect image generation in the second stage. Unfortunately, the ground-truth pose $P_{t_{i}}$ does not exist during inference for pose correct purposes - it is only available for training. Therefore, it is necessary to teach training to properly handle abnormal poses with the guidance of ground-truth pose, in order to generalize to testing scenarios.

Through the observation of heatmaps of those abnormal poses, we find that they are often due to the small differences in corresponding key point responses, which will not contribute to large loss and thus incur small back-propagation gradients. As a matter of fact, there is no need to push the limit of the pose generation accuracy by PSGAN since the small errors should not affect how people interpret pose globally. Considering the pose prediction difference is inevitably noisy at the input layer or low-level feature layer, we propose to enforce the semantic consistency between abnormal poses and the ground-truth at the high-level feature layer.

Fig.~\ref{fig:APPGAN} shows our Semantic Consistent GAN that encapsulates this idea. We share weights in the last convolutional layer of two pose encoder networks (the yellow block), aiming to impose semantic consistent in the high-level feature space. Moreover, we generate video frames from both predicted pose and ground-truth pose to gain tolerance to pose noise. A new discriminator $D_{which}$ is used to distinguish which pose the generated video frame is conditioned on. We further utilize the $L1$ reconstruction loss to stabilize the training process.

\subsubsection{Full objective function.}

As shown in Fig. \ref{fig:APPGAN}, our final objective is to generate video frames from two pose streams and keep their semantic consistency in an adversarial way. Specifically, $G_{1}$ generates the image $I_{t_{i}|gen}$ at time $t_{i}$ conditioned on input image $I_{t_{0}}$ and the pose $\hat{P}_{t_{i}}$ generated by PSGAN. $G_{2}$ generates $I_{t_{i}|gt}$ in the same way but uses the ground-truth pose for image generation.
\begin{align}
\label{Gone}
& G_{1}(I_{t_{0}}, \hat{P}_{t_{i}}) \Rightarrow I_{t_{i}|gen}, \\
\label{Gtwo}
& G_{2}(I_{t_{0}}, {P_{t_{i}})} \Rightarrow I_{t_{i}|gt}.
\end{align} 
There are three discriminators defined as follows: $D_{1}$ and $D_{2}$ aim to distinguish between real image and fake when using predicted pose and ground-truth pose respectively; $D_{which}$ aims to judge which pose the generated image is conditioned on. Then we easily arrive at the full objective function for our model training as follows:
\begin{align}
\mathcal{L}_{D_{which}} &= \mathbb{E}[\log(D_{which}(I_{t_{i}|gt}))] + \mathbb{E}[{\log(1 - D_{which}(I_{t_{i}|gen}))}], \\
\mathcal{L}_{D_{1}} &= \mathbb{E}[\log(D_{1}(I_{t_{i}}))] + \mathbb{E}[{\log(1 - D_{1}(I_{t_{i}|gen}))}], \\
\mathcal{L}_{D_{2}} &= \mathbb{E}[\log(D_{2}(I_{t_{i}}))] + \mathbb{E}[{\log(1 - D_{2}(I_{t_{i}|gt}))}], \\
\mathcal{L}_{G_{1}} &= \mathbb{E}[{\log(D_{1}(I_{t_{i}|gen}))}] + \mathbb{E}[{\log(D_{which}(I_{t_{i}|gen}))}], \\
\label{dwhich}
\mathcal{L}_{G_{2}} &= \mathbb{E}[{\log(D_{2}(I_{t_{i}|gt}))}].
\end{align} 
Since the ground-truth pose guided images $I_{t_{i}|gt}$ is \emph{real} for $D_{which}$, the gradient of $D_{which}$ is not propagated back to $G_{2}$ in Eq. (\ref{dwhich}).

\subsection{Implementation Details}

For our detailed network architecture, all of the generators $(G,G_{1},G_{2})$ apply 4 convolution layers with a kernel size of 4 and the stride of 2 for downsampling. In the decoding step of stage one, transposed convolution layers with stride of 2 are adopted for upsampling, while normal convolutions layers together with interpolation operations take place of transposed convolution layers in the second stage. The feature map of the red block in Fig.~\ref{fig:PSGAN} is extended with a time dimension ($C \times W \times H \Rightarrow C \times 1 \times W \times H$) for the decoder of PSGAN. The discriminators $(D, D_{1}, D_{2}, D_{which})$ are PatchGANs~\cite{isola2017image} to classify whether local image patches are real or fake. Besides, ReLU \cite{nair2010rectified} serves as the activation function after each layer and the instance normalization \cite{IN} is used in all networks. Several residual blocks~\cite{he2016deep} are leveraged to encode the concatenated feature representations jointly. For the last layer we apply tanh activation function. In addition, we use standard GRU in our PSGAN, without further investigating how the different structures of LSTM can improve pose sequence generation.

We implement all models using PyTorch, and use an ADAM \cite{kingma2014adam} optimizer with a learning rate of $0.001$ in all experiments. The batch size in stage one is $64$, and 128 in the second stage. All reconstruction loss weights are empirically set to $10$. The scaling factor $\alpha$ in Eq.(\ref{maskl1}) is chosen from 0 to 100, which only affects the convergence speed. We empirically set the scaling factor as 10 and 20 on the human action and facial expression dataset respectively. The PSGAN is trained to generate pose sequences. In the second stage, both the generated and ground-truth poses are utilized to train the SCGAN to learn robust handling of noisy poses. Only the generated pose is fed into SCGAN for inference.

\section{Experiments}

In this section, we present video generation results on both human action and facial datasets. Qualitative and quantitative comparisons are provided to show our superiority over other baselines and state-of-the-arts. User study is also conducted with a total of 50 volunteers to support our improvements. Ablation study for our two generation stages (for pose and video) is further included to show their efficacy.

\subsection{Datasets}
Our experiments are conducted not only on the human action dataset but also on facial expression dataset, where facial landmarks act as the pose to guide the generation of facial expression videos. Accordingly, we collected the Human Action Dataset and Human Facial Dataset as detailed below. For all experiments, the RGB images are scaled to $128 \times 128$ pixels while pose images are scaled to $64 \times 64$ pixels. 

\begin{itemize}
\item [$\bullet$] Human Action Dataset comes from the UCF101~\cite{soomro2012ucf101} and Weizmann Action database~\cite{blank2005actions}, including 198848 video frames of 90 persons performing 22 actions. Human pose is extracted by the method in~\cite{cao2017realtime} with 18 key points.
\item [$\bullet$] Human Facial Dataset is from the $CK+$ dataset~\cite{lucey2010extended}. We consider $6$ facial expressions: \emph{angry}, \emph{disgust}, \emph{fear}, \emph{happy}, \emph{sadness} and \emph{surprise}, corresponding to 60 persons and 60000 frames. The facial pose is annotated with 68 key points. 
\end{itemize}

\subsection{Evaluation of Pose Generation}
\begin{figure}[t]
\centering
\includegraphics[width=1\textwidth]{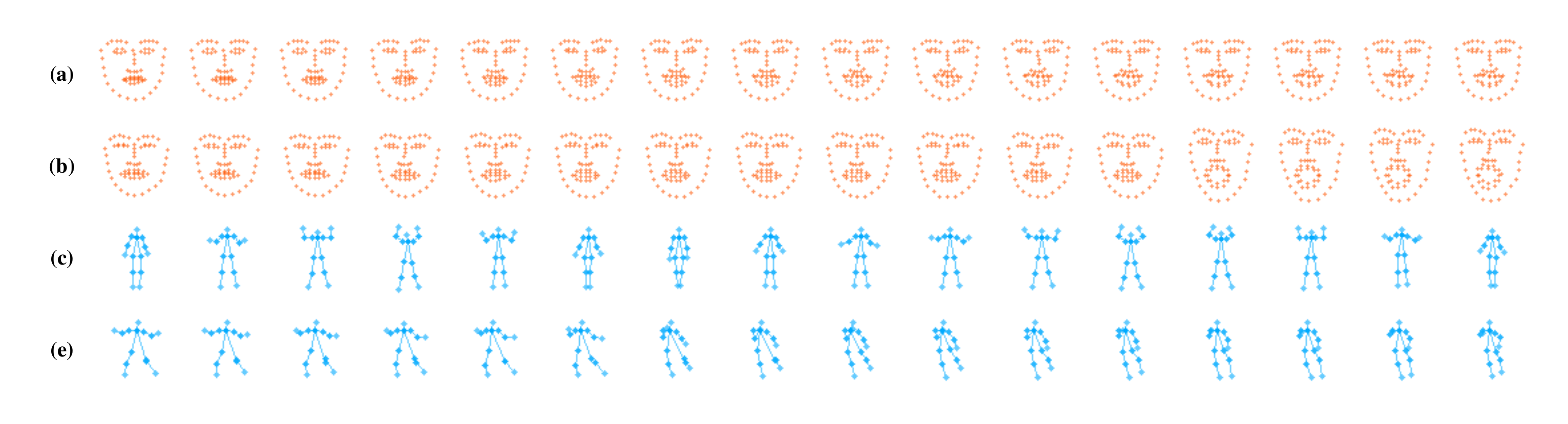}
\caption{Example pose sequences generated by our PSGAN with class labels of \emph{Happy} (a), \emph{Surprise} (b), \emph{Wave} (c) and \emph{TaiChi} (d), respecitvely}
\label{fig:POSE}

\end{figure}
\subsubsection{Qualitative evaluation.}
As mentioned in Section~\ref{Ill}, our PSGAN focuses on the generation of various pose motions. For qualitative comparisons, we follow the post-processing step in~\cite{cao2017realtime} to locate the maximum response region in each pose heatmap. Note such pose-processing is only used for visualization purposes. Fig.~\ref{fig:POSE} shows some examples of the generated pose sequences for both human face and body. We can see that the pose sequences change in a smooth and typical way under each action scenario. 

\subsubsection{Quantitative comparison.}
Recall that our final video generator is tolerant to the tiny pose difference in stage one. Therefore, we measure the quality of generated pose sequences by calculating the average pairwise $L2$ distance rather than Euclidean norm between the generated and the ground-truth poses. Smaller distance indicates better pose quality.

We compare three of our PSGAN variants: 1) PSGAN trained with the $L1$-norm loss rather than adversarial loss, 2) PSGAN trained without the LSTM module, and 3) the full PSGAN model with a GRU module. Table~\ref{table:AL} indicates that it is better to train our pose generator with the adversarial loss than with the simple $L1$-norm loss. Also important is the temporal modeling by the GRU or LSTM module that improves the quality of pose sequences.

\begin{table}[t]
\begin{minipage}{0.5\textwidth}
\centering
\caption{Quantitative comparison of \protect\\ pose generation baselines}
\label{table:AL}
\resizebox{0.8\textwidth}{!}{
\begin{tabular}{lcc}
\hline\noalign{\smallskip}
Average $L2$ & Action &Facial Exp.\\
\noalign{\smallskip}
\hline
\noalign{\smallskip}
Ground-truth & 0 & 0\\
PSGAN-$L1$ & 0.0124 & 0.0078\\
PSGAN w/o LSTM & 0.0072 & 0.0062\\
PSGAN &{\bf 0.0064} & {\bf 0.0051}\\
\hline
\end{tabular}
}
\end{minipage}
\begin{minipage}{0.5\textwidth}
\centering
\caption{User study of pose generation baselines on human action dataset}
\label{table:US}
\resizebox{0.8\textwidth}{!}{
\begin{tabular}{lcccc}
\hline\noalign{\smallskip}
Distribution of ranks & 1 & 2 & 3 & 4 \\
\noalign{\smallskip}
\hline
\noalign{\smallskip}
Ground-truth & 0.38 & 0.36 & 0.12 & 0.14 \\
PSGAN-$L1$ & 0.09 & 0.08 & 0.32 & 0.51 \\
PSGAN w/o LSTM & 0.21 & 0.16 & 0.43 & 0.20 \\
PSGAN & 0.32 & 0.40 & 0.13 & 0.15 \\
\hline
\end{tabular}
}
\end{minipage}
\end{table}

\subsubsection{User study.}
Table~\ref{table:US} includes the user study results for our three PSGAN variants on human action dataset. For each variant, we generate 25 pose sequences with 20 actions and the time step of 32. All generated pose sequences are shown to 50 users in a random order. Users are then asked to rank the baselines based on their quality from 1 to 4 (best to worst). The distribution for the ranks of each baseline is calculated for comparison. As shown in Table ~\ref{table:US}, our full PSGAN model has the highest chance to rank top. While the variants of PSGAN w/o LSTM and PSGAN-$L1$ tend to rank lower, indicating the importance of temporal and adversarial pose modeling again.

\subsection{Evaluation of Video Generation}

\subsubsection{Qualitative comparison.}
\begin{figure}[t]
\centering
\includegraphics[width=0.8\textwidth]{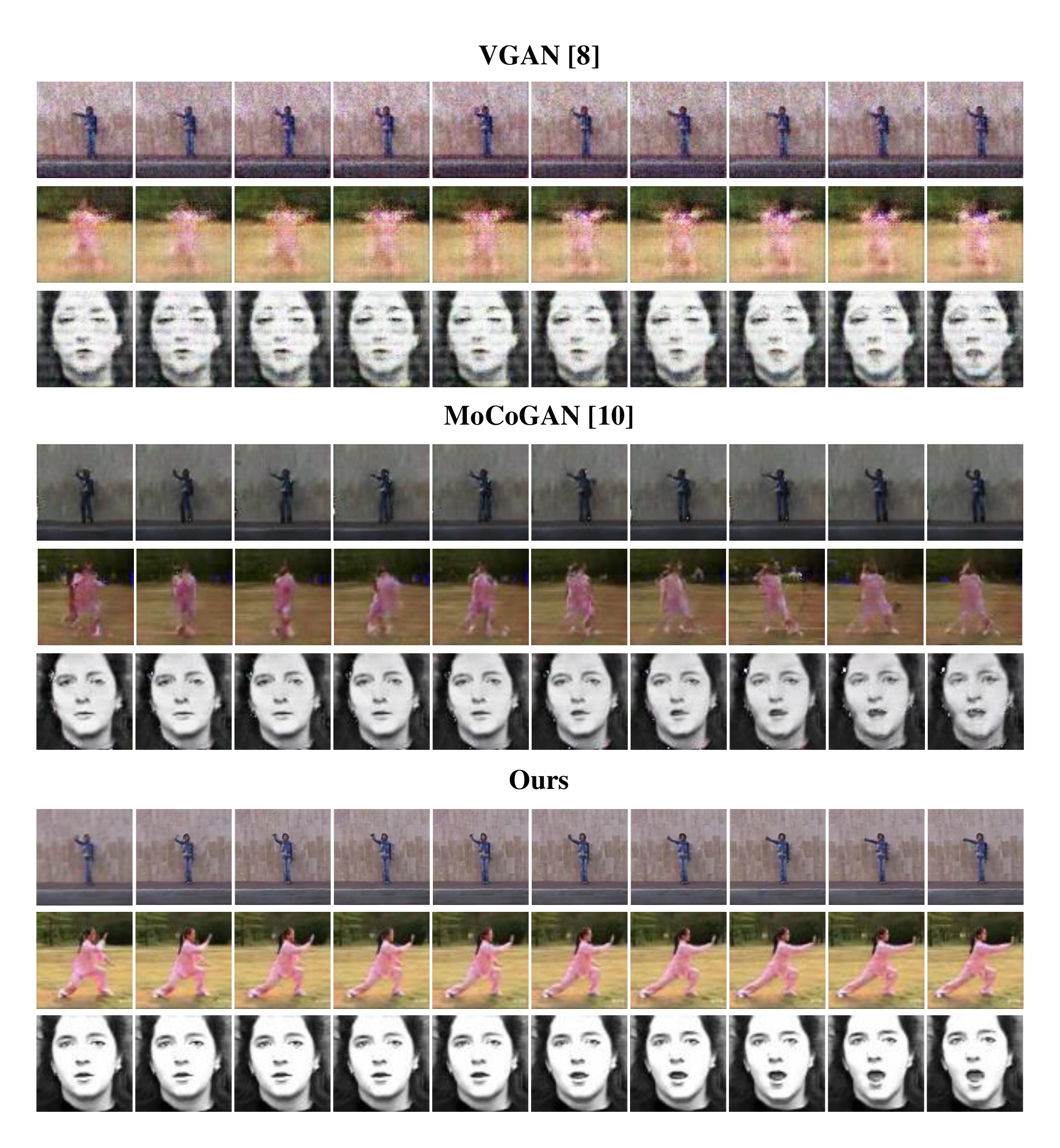}
\caption{Generated video frames for example action and facial expression classes \emph{Wave}, \emph{Taichi} and \emph{Superised} by VGAN~\cite{vondrick2016generating}, MoCoGAN~\cite{tulyakov2017mocogan} and our method}
\label{fig:US}
\end{figure}

Given the generated pose sequence in the first stage and the input image, our SCGAN is responsible for the generation of photo-realistic video frames. We mainly compare our method with state-of-the-art video generation methods VGAN~\cite{vondrick2016generating} and MoCoGAN~\cite{tulyakov2017mocogan}. They are trained on the same human action and facial datasets with hyper-parameters tuned to their best performance. The visual results for example action and facial expression classes \emph{Wave}, \emph{Taichi} and \emph{Superised} are shown in Fig.~\ref{fig:US}.

It is clear that our method generates much sharper and more realistic video frames than VGAN and MoCoGAN. For the simple action \emph{Wave}, our method performs better or equally well with the strong competitors. For the difficult action \emph{TaiChi} with complex motion patterns, our advantage is evident - the pose dynamics are accurately captured and rendered to visually pleasing images. This confirms the necessity of our pose-guided video generation which benefits from explicit pose motion modeling rather than using a noise vector in VGAN and MoCoGAN. Our supplementary material provides more visual results.

\begin{table}[b]
\begin{minipage}{0.48\textwidth}
\centering
\caption{Comparison of IS for video generation baselines}
\label{table:IS}
\resizebox{0.98\textwidth}{!}{
\begin{tabular}{lcc}
\hline\noalign{\smallskip}
IS & Action &Facial Exp.\\
\noalign{\smallskip}
\hline
\noalign{\smallskip}
VGAN~\cite{vondrick2016generating} & 2.73 $\pm$ 0.21 & 1.68 $\pm$ 0.17\\
MoCoGAN~\cite{tulyakov2017mocogan} & 4.02 $\pm$ 0.27 & 1.83 $\pm$ 0.08\\
Ours & {\bf 5.70 $\pm$ 0.19} & {\bf 1.92 $\pm$ 0.12 }\\
\hline
\end{tabular}
}
\end{minipage}
\begin{minipage}{0.45\textwidth}
\centering
\caption{User Study for video generation baselines}
\label{table:USV}
\begin{tabular}{lcc}
\hline\noalign{\smallskip}
Winning percentage & Action & Facial Exp.\\
\noalign{\smallskip}
\hline
\noalign{\smallskip}
Ours/MCGAN~\cite{tulyakov2017mocogan} & 0.83/0.17 & 0.86/0.14  \\
Ours/VGAN~\cite{vondrick2016generating} & 0.88/0.12 & 0.93/0.07  \\
\hline
\end{tabular}
\end{minipage}
\end{table}

\subsubsection{Quantitative comparison.}
Table.~\ref{table:IS} shows the measures of Inception Score \cite{salimans2016improved} (IS) (and its variance) for different methods. Larger IS value indicates better performance. Such quantitative results are in line with our visual evaluations where our method outperforms others by a large margin.

\subsubsection{User study.}
We further conduct a user study where each method generates 50 videos for comparisons. Results are provided to users in pairs and in random order. The users are then asked to select the winner (looks more realistic) from the paired methods and we calculate the winning percentage for each method. Table~\ref{table:USV} demonstrates that most of the time users would choose our method as the winner over MoCoGAN and VGAN.

\subsubsection{Controlled generation results.}
\begin{figure}[t]
\centering
\includegraphics[width=1\textwidth]{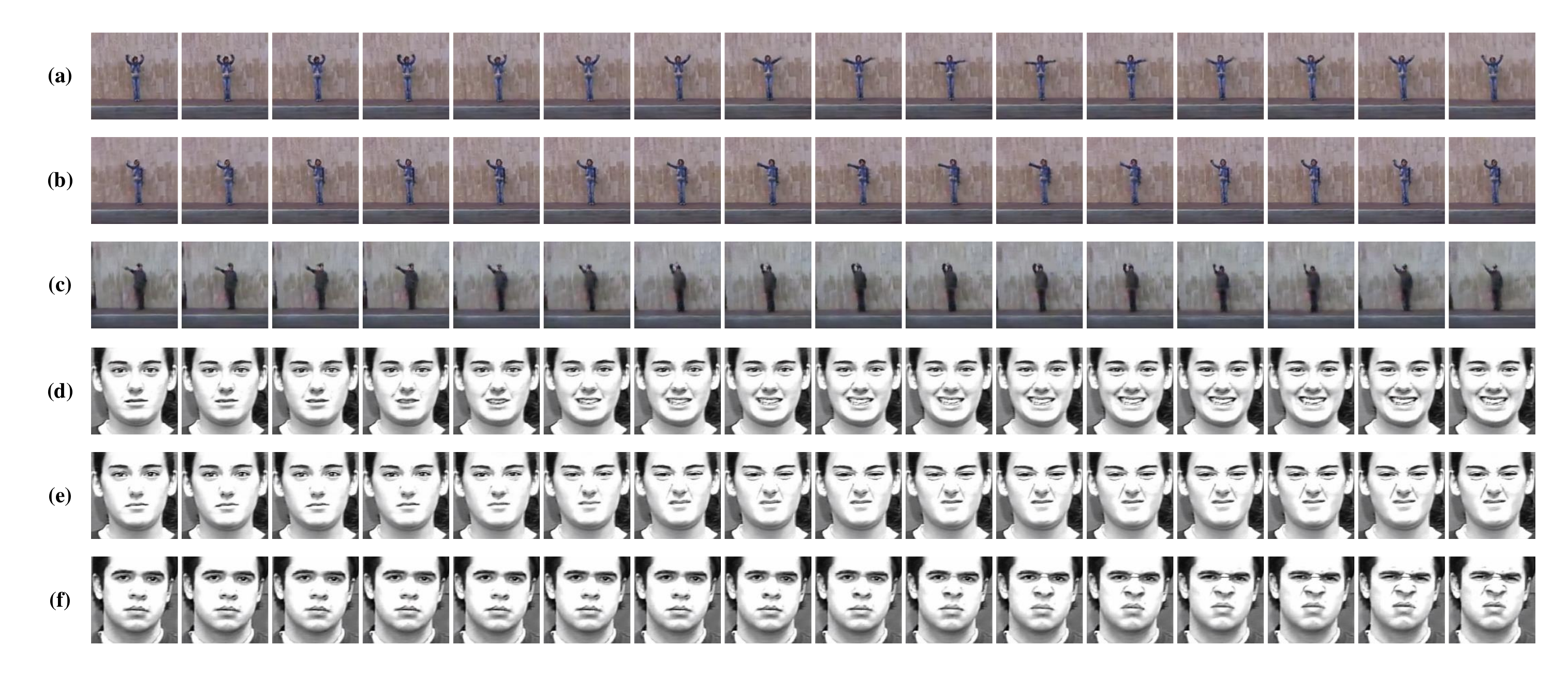}
\caption{Controlled video generation with pose and appearance: different poses for the same human (a-b for body, d-e for face), and same pose on different humans (b-c for body, e-f for face)}
\label{fig:IMG}
\end{figure}

Fig.~\ref{fig:IMG} validates our capability of explicit pose modeling and good generalization ability by controlled tests: generate different action videos with a fixed human appearance, and generate videos with a fixed action for different humans. Successes are found for both human action (a-c) and facial expression (d-f) cases, showing the benefits of separate modeling for pose and appearance.

\begin{table}[b]
\begin{center}
\setlength{\tabcolsep}{4mm}
\caption{SSIM and LPIPS measures for our training alternatives}
\label{table:SSIM}
\begin{tabular}{lcc}
\hline\noalign{\smallskip}
SSIM/LPIPS & Action &Facial Exp.\\
\noalign{\smallskip}
\hline
\noalign{\smallskip}
Static & 0.66/0.063 & 0.77/0.025 \\
SCGAN-gen & 0.73/0.083 & 0.89/0.038\\
SCGAN-gt & 0.89/0.040 & 0.92/0.024 \\
\hline
SCGAN    & 0.87/0.041 & 0.91/0.026\\
\hline
\end{tabular}
\end{center}
\end{table}

\subsection{Ablation Study}

One major feature of our human video generator is its reliance on generated pose $\hat{P}_{t_{i}}$. It is worth noting that there is no ground-truth pose $P_{t_{i}}$ during inference. Only for training, we use the available $P_{t_{i}}$ to enforce the semantic consistency with respect to generated pose $\hat{P}_{t_{i}}$. To highlight the impact of the semantic consistency constraint, we compare several training alternatives as follows:
\begin{itemize}
\item Static generator: video generation by repeating the first frame
\item SCGAN-gen: video generation guided by generated pose $\hat{P}_{t_{i}}$ only.
\item SCGAN-gt: video generation guided by ground-truth pose $P_{t_{i}}$ only.
\item SCGAN: video generation guided by both $\hat{P}_{t_{i}}$ and $P_{t_{i}}$ as shown in Fig.~\ref{fig:APPGAN}.
\end{itemize}

The baseline of static generator simply constructs a video by repeating the first frame thus involves no predictions. It acts as a performance lower bound here. Sometimes its performance can be not too bad in those short videos or videos with little change (generating a static video will not be heavily penalized in these cases).

\begin{figure}[t]
\centering
\includegraphics[width=1\textwidth]{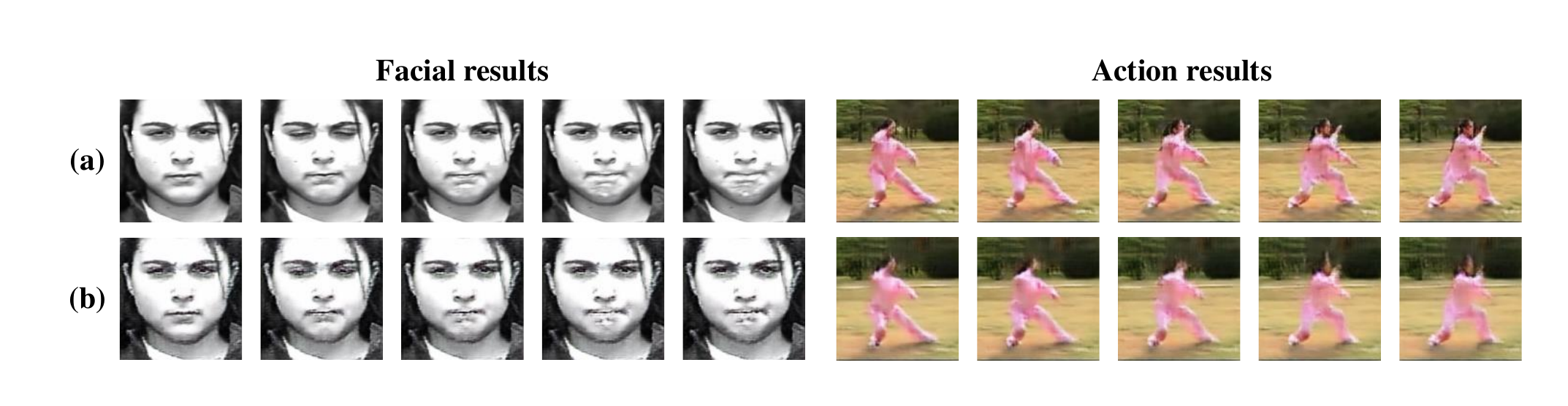}
\caption{Visual comparisons of SCGAN (a) and SCGAN-gen (b) on facial expression and human action datasets}
\label{fig:IMPROVE}
\end{figure}

Fig.~\ref{fig:IMPROVE} visually compares the baselines of SCGAN-gen and SCGAN. The full SCGAN model can generate sharper and more photo-realistic results, especially in the {\bf mouth} (facial expression) and {\bf waist} (human action) regions. This suggests the efficacy of enforcing semantic consistency between the generated and ground-truth poses. Using the generated pose only can be noisy and thus hinder the final video quality.

We also evaluate performance by computing the SSIM (structural similarity index measure) \cite{wang2004image} and LPIPS (Learned Perceptual Image Patch Similarity) scores~\cite{zhang2018unreasonable}. The SSIM score focuses on structural similarity between the generated image and ground-truth, while the LPIPS score cares more about perceptual similarity. Higher SSIM score and smaller LPIPS score indicates better performance.

Table ~\ref{table:SSIM} shows that SCGAN indeed outperforms SCGAN-gen quantitatively and stands close to the SCGAN-gt using ground-truth pose. The semantic consistency constraint plays a key role in this improvement, because it can alleviate the influence of abnormal poses during the pose guided image generation process. When compared to the static video generator, our method outperforms by generating a variety of motion patterns.

\section{Conclusion and Future Work}

This paper presents a novel method to generate human videos in a disentangled way. We show the important role of human pose for this task, and propose a pose-guided method to generate realistic human videos in two stages. Quantitative and qualitative results on human action and face datasets demonstrate the superiority of our method, which is also shown to be able to manipulate human pose and appearance explicitly. Currently, our method is limited to cropped human or face images since the detectors are missing. In the future, we will integrate detectors as an automatic pre-processing step which will enable multi-person video generation.

\noindent \textbf{Acknowledgement. } This work is partially supported by the Big Data Collaboration Research grant from SenseTime Group (CUHK Agreement No. TS1610626). 

\bibliographystyle{splncs}
\bibliography{egbib}
\end{document}